\documentclass[lettersize,journal]{IEEEtran}
\usepackage{amsmath,amsfonts}
\usepackage{algorithmic}
\usepackage{algorithm}
\usepackage{array}
\usepackage[caption=false,font=normalsize,labelfont=sf,textfont=sf]{subfig}
\usepackage{textcomp}
\usepackage{stfloats}
\usepackage{url}
\usepackage{verbatim}
\usepackage{graphicx}
\usepackage{cite}
\usepackage{multirow}
\usepackage[normalem]{ulem}
\useunder{\uline}{\ul}{}
\begin{document}

\title{Spatiotemporal Multi-scale Bilateral Motion Network for Gait Recognition}

\author{Xinnan Ding, Shan Du, Yu Zhang, and Kejun Wang
\thanks{Manuscript received   xx; revised      xx. The work is supported by National Natural Science Foundation of China (61573114) and the program of China Scholarships Council (202006680049). (The corresponding author is Kejun Wang.)}
\thanks{Xinnan Ding is with College of Intelligent Systems Science and Engineering, Harbin Engineering University, China. (email: dingxinnan@hrbeu.edu.cn)}
\thanks{Shan Du is with Department of Computer Science, Mathematics, Physics and Statistics, University of British Columbia Okanagan, Kelowna, Canada. (email: shan.du@ubc.ca)}
\thanks{Yu Zhang is with School of Mathematics, Harbin Institute of Technology, Harbin, China. (email: math\_zhyu@163.com)}
\thanks{Kejun Wang is with College of Intelligent Systems Science and Engineering, Harbin Engineering University, and Beijing Institute of Technology Zhuhai, Zhuhai, China. (email: wangkejun@hrbeu.edu.cn)}}


\maketitle

\begin{abstract}
The critical goal of gait recognition is to acquire the inter-frame walking habit representation from the gait sequences. The relations between frames, however, have not received adequate attention in comparison to the intra-frame features. In this paper, motivated by optical flow, the bilateral motion-oriented features are proposed, which can allow the classic convolutional structure to have the capability to directly portray gait movement patterns at the feature level. Based on such features, we develop a set of multi-scale temporal representations that force the motion context to be richly described at various levels of temporal resolution. Furthermore, a correction block is devised to eliminate the segmentation noise of silhouettes for getting more precise gait information. Subsequently, the temporal feature set and the spatial features are combined to comprehensively characterize gait processes. Extensive experiments are conducted on CASIA-B and OU-MVLP datasets, and the results achieve an outstanding identification performance, which has demonstrated the effectiveness of the proposed approach. 
\end{abstract}


\section{Introduction}
\IEEEPARstart{V}{ideo-based} gait recognition aims at identifying individuals by monitoring and analyzing their walking habits. As a kind of distinct and promising biometric characteristics, gait recognition has several inherent strengths. Concretely, like faces, gait recognition obtains the identities through surveillance videos directly, which does not always necessitate the cooperation of subjects. Similar to signatures, behavioral habits are very difficult to conceal or imitate. These properties give gait recognition potential, exclusive, and notable superiority in access control, criminal investigation, and border security.

There are, however, numerous challenges in real-world gait recognition \cite{RN78}. Specifically, the dominant concerns are how to lessen the influence of external factors and deal with video data. For one thing, the subject has different walking directions, clothes, or accessories in the actual situation, which bring disturbances to the extracted gait features \cite{RN63,RN76}. Although some exterior traits, like clothing, are often used as the key attribute for cross-camera pedestrian re-identification (Re-ID) tasks \cite{RN85}, it is not unique or unchangeable, which results in mistakes, especially for those who deliberately hide their identity, such as criminals. Recently, a growing number of researchers have focused on cross-dressing Re-ID to reduce the reliance on clothing \cite{RN83,RN84}. Accordingly, retrieving the information of gait habits and removing the side effect of external variables is of top priority. For another thing, gait modality exists in consecutive frames, instead of an image. Due to the complexity of gait sequences, capturing gait habit features from a series of frames is the remaining obstacle to overcome. The relations between these continuous frames should be analyzed, distilled, and integrated to represent the gait features\cite{RN31}. Therefore, understanding and learning the whole video to get discriminative features for identification is the crucial task of gait recognition.

From the perspective of processing videos, existing studies can be classified into two main categories: set-based and sequence-based. Creating a template (an image) that can embody a gait video, like gait energy images (GEIs) \cite{RN35}, is a classic type of set-based methods \cite{RN37,RN43,RN82}. Another influential type of set-based researches extracts features from each gait shuffled silhouette\cite{RN1,RN64,RN33}. These approaches consider the video as a random collection of images. Although they make the task more accessible and less expensive to compute, the cross-frame features conveying the habit traits are ignored. Sequence-based methods, on the other hand, pay more attention to inter-frame movement information. For example, 3D convolutional neural networks (CNNs) can directly capture gait features from the sequence\cite{RN72,RN41}; abstracted skeleton information can model the human body reflecting movement features\cite{RN86,RN39}; and some elaborate units in CNNs can be designed to learn and integrate multi-frame information\cite{RN56}. However, such attempts often depend on a special and sophisticated sequential structure.

In order to address this issue, we present the spatiotemporal multi-scale bilateral motion (SMBM) network. SBMB can allow CNNs to understand motion information, the scheme of which is illustrated in Fig. \ref{fig1}. First, we design a bilateral motion block, which is a 2D-CNN structure to obtain a kind of feature-level optical-flow-like representation. Bilateral motion-oriented features are able to describe a sophisticated motion context with solid theoretical interpretability. Second, a set of these features is developed at several coarse-to-fine temporal resolutions, which covers a different number of frames, thereby getting more specific gait habit features.  This generalization method can be implemented by stacked blocks or the high-order gradient method. Third, a correction block is designed to lower the disturbance produced by inaccurate segmentation. Therefore, the contributions of this work are as follows:

•	The bilateral motion block is proposed to explore the motion descriptions in 2D-CNNs. Particularly, a novel kind of gait features inspired by the optical flow algorithm, the bilateral motion-oriented features, can intuitively reflect gait habits in this block. 

•	The features are then generalized to a group of multi-scale temporal representations to comprehensively depict the motion from multiple resolutions, where an embedded generalizing approach employing high order gradients can refine the features.

•	Gait silhouettes are often obtained inaccurately due to the long distance, so we construct a correction block to distinguish dynamic information from segmentation noise, whose adverse impact would be restricted.

•	A wide range of experiments are carried out on two popular public benchmarks, CASIA-B and OU-MVLP. The results of our method outperform other related algorithms and verify its robustness and effectiveness.

\section{Related work}
Deep-learning-based gait recognition researches are dedicated to capturing the invariant features that would not be affected by exterior factors. Like classic approaches, such as human 3D models\cite{RN74,RN75}, transform models \cite{RN81}, and characteristics analysis \cite{RN34,RN77,RN80}, the gait recognition based on deep learning can be divided into two categories: the appearance-based methods and the motion-based methods.
\subsection{Appearance-based gait recognition}
Deep learning makes extracting gait traits without modeling possible. In the beginning, CNNs are applied to learn the distinctive features directly for gait recognition \cite{RN52, RN11,RN21}. For example, Wu et al. \cite{RN2} designed a CNN network to learn GEI, and Chao et al. \cite{RN1} constructed a specialized CNN called GaitSet to capture the gait features from a set \cite{RN33}, improving the rank-1 accuracy to a new grade. However, these methods are not very robust to complex scenarios, and altering views would exert a marked impact on the human appearance. To this end, some studies seek to generate or transform normalized images or features \cite{RN25}, where auto-encoders and generative adversarial networks are attempted \cite{RN81,RN68,RN3, RN12}. Besides, carrying and clothing conditions are unfavorable to extract the invariant features. There are a lot of researches employing the partition algorithm to resolve this issue \cite{RN70}, since bags or jackets only affect a part of the body. In particular, Fan et al. \cite{RN15} proposed the focal convolution that can straightforwardly capture local features. Lin et al. \cite{RN72} created the global and local extractor for features in parts. Yao et al. \cite{RN57} assembled the local features and body skeleton to portray the gait appearance. Furthermore,   the correlations between parts were investigated to better model the human body by RNN and LSTM \cite{RN10, RN58, RN71}. Appearance-based approaches are more focused on the single frame-level elements, which play a vital role in gait recognition, but the video-level features are not supposed to be overlooked.
\subsection{Motion-based gait recognition}
There are mainly two kinds of ways to explore the motion context of gait recognition. The first type of methods adopts the human skeleton to depict the abstract pose and movement reflecting the motion dynamics \cite{RN38}, as the pose estimation approaches can provide the body key-point information readily \cite{RN44,RN45}. Liao et al. \cite{RN40} applied 3D pose estimation to construct the human model, and RNN or LSTM were deployed to integrate gait features from the model. Li et al. \cite{RN67} combined the graph convolutional network and multi-scale skeletons to extract the temporal features. The second kind of strategies employs  networks that can understand temporal dynamic behavior. Wolf et al. \cite{RN41} first employed the 3D CNN for gait recognition to acquire the distinctive gait features. Then, Lin et al. \cite{RN59,RN72}  developed the 3D CNN to capture the spatiotemporal features from video with unfixed length. Zhang et al. \cite{RN8} integrated the sequence-level features from each RGB frame by the LSTM network. Ding et al. \cite{RN56} constructed a frame difference block that forces the CNN to learn the motion context. Fan et al. \cite{RN15} designed a sequence attention unit to weigh each frame in a short range to gather information from the entire video. Accordingly, capturing motion information contributes to gait identification, but there is a lack of useful and reasonable representations that can be embedded in neural networks.

\section{Proposed Method}\label{3}
\subsection{Overview}
\begin{figure*}[!t]
\centering
\includegraphics[width=18cm]{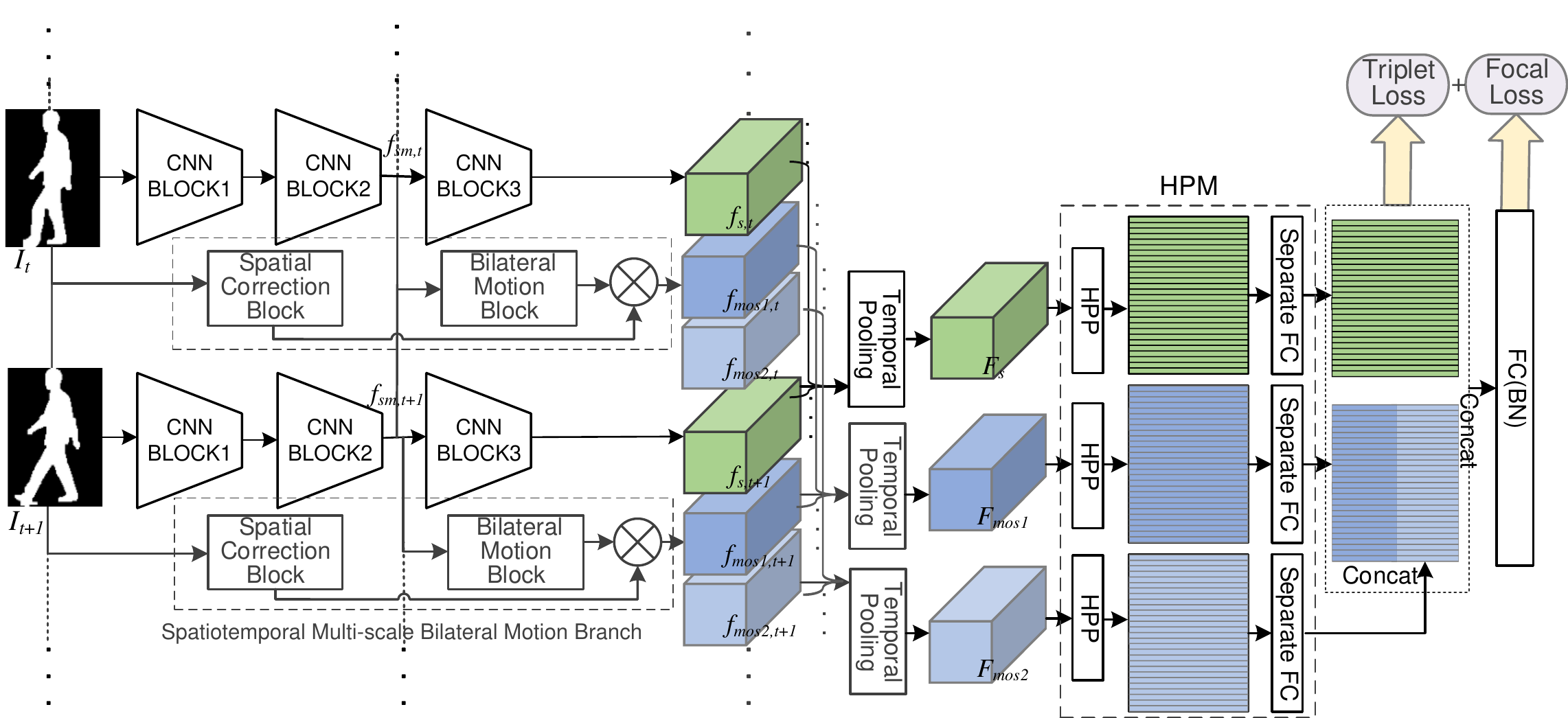}
\caption{The framework of SMBM network. HPP is horizontal pyramid pooling and HPM represents horizontal pyramid mapping. Spatial correction block is used for evaluating the accuracy of segmentation and can mitigate the effect of inaccurate parts. Bilateral motion block generates multi-scale feature-level motion representation.}
\label{fig1}
\end{figure*}
As illustrated in Fig. \ref{fig1}, the SMBM network has a two-branch architecture, where the spatiotemporal multi-scale bilateral motion branch is added to the classic gait recognition structure\cite{RN1}. By this means, the traditional CNN can learn and capture sequential information in a gait video without RNNs or LSTM. A sequence including $n$ frames $\tilde{I}=\{{{I}_{1}},{{I}_{2}},\ldots ,{{I}_{n}}\}$ is fed into the network. For one branch, the backbone consisting of three CNN blocks is applied to extract frame-level features ${{\tilde{f}}_{s}}=\{{{f}_{s,1}},{{f}_{s,2}},\ldots ,{{f}_{s,n}}\}$ of each silhouette. The operation can be formulated as:
\begin{equation}
\label{eq1}
{{f}_{s,t}}=C({{I}_{t}})
\end{equation}
where ${{f}_{s,t}}$ denotes the obtained features of the $t$ frame ${{I}_{t}}$, $C(\cdot)$ is the CNN network. As ${{f}_{s,t}}$ comes from every independent frame, it reflects static appearance information. For another branch, there are two major components: a correction block and a bilateral motion block. Because segmentation noise confuses the network to distinguish dynamic features between frames, it is necessary for the correction block to filter out this type of noise. This block learns from the neighbors of the target frame of the whole raw video $\tilde{I}$, so that it can estimate segmentation faults to calibrate the exacted movement. For the bilateral motion block, the input is the adjacent frame-level features generated by the middle layer of the first branch, since focusing on a local temporal range is more capable of capturing gait habit. This block can extract motion-oriented features from the abstract tensor. Besides, a group of features is generalized to ${i}$ scales as refinement, i.e., ${{\tilde{f}}_{mo,t}}=\{{{f}_{mos1,t}},{{f}_{mos2,t}},\ldots ,{{f}_{mosi,t}}\}$, and it can be represented as:
\begin{equation}
\label{eq2}
{{\tilde{f}}_{mo,t}}=Cor(\tilde{I})\cdot BM({{f}_{sm,t-1}},{{f}_{sm,t}},{{f}_{sm,t+1}})
\end{equation}
where $Cor(\cdot)$ is defined as the function indicating the correction block,  $BM(\cdot)$ is the bilateral motion block, and ${{f}_{sm,t}}$ denotes the static middle-level features from another branch. As ${{f}_{mos1,t}},{{f}_{mos2,t}},\ldots ,{{f}_{mosi,t}}$ lie at different levels of resolution and span different temporal ranges, a set of ${{f}_{sm}}$ with different numbers of frames is fed into the block based on the setting of $i$ scales. The number would grow as $i$ increases. Eq. \ref{eq2} is the case of two scales as an example, as shown in Fig. \ref{fig1}, so the input is ${{f}_{sm,t-1}},{{f}_{sm,t}},{{f}_{sm,t+1}}$. The specific scheme of the bilateral motion block will be illustrated in Section \ref{3.2}.

Temporal pooling and horizontal pyramid mapping (HPM) are proved as two widely applied and effective units for gait recognition\cite{RN1,RN15}. In this research, temporal pooling is independently deployed for ${{\tilde{f}}_{s}}$ and ${{\tilde{f}}_{mo}}$ at each scale to learn the most representative appearance information. Then the HPM is exploited, where the horizontal pyramid pooling (HPP) forces the feature map to be cut into multiple strips, and the global max and mean pooling to be combined to integrate local and global information\cite{RN91}. Separate fully connect (FC) layers in HPM can map the features to a more discriminative space. Next, the motion-oriented feature map at different resolutions is concatenated in the horizontal axis, and the static features and the motion features from the two branches are also concatenated in the vertical axis. In this case, HPMs enforce these two kinds of features to be mapped into the same size. In addition, the triplet loss and the focal loss are adopted jointly here, which is the first to apply these two losses in gait recognition.  Gait recognition is a special task with properties on both metric learning and classification. The triplet loss is appropriate for metric learning and representational learning\cite{RN19}, and the focal loss is a good measure of the probability distribution of classification results. Compared to the cross-entropy (CE) loss, the focal loss can weaken the importance of simple negative samples in the training\cite{RN87}.

\subsection{Bilateral motion-oriented block}\label{3.2}
Motion depiction in videos is a tricky issue to address. Frame difference and optical flow are two types of major traditional image processing algorithms\cite{RN90}. In deep learning, some particular structures, such as 3D-CNN, RNN, and LSTM, are put forward and achieve remarkable results on temporal information learning tasks. They can learn via the memory of a sequence of data, and some similar methods are also employed in gait recognition\cite{RN59,RN72,RN41}. However, these attempts are usually complicated, poorly explainable, and dependent on data and computational resources. To this end, we consider adopting traditional motion representation at the feature level for gait recognition. Our previous work tends to explore the feature-level frame difference deployed in the CNN network\cite{RN56}, but differences are disturbed by segmentation accuracy and are too simple to adequately convey motion. Hence, inspired by the optical flow algorithm, we apply bilateral motion-oriented features in the network, and develop such features at multiple temporal scales. Furthermore, a correction block is designed to diminish the impact of the wrong segmentation.

Optical flow is a concept to describe the observed instantaneous velocity of the pixel motion. In video processing, the optical flow describes the motion information via the change of pixels in the time domain. Some optical-flow-like features adopted in action recognition achieve excellent performances\cite{RN88,RN89}, which imply that satisfactory results in a similar task, gait recognition, are conceivable. Motivated by the optical flow algorithms, the motion-oriented features can be defined under the brightness constant constraint\cite{RN90}. For frames, $I_t$ and ${{I}_{t+\Delta t}}$ in a gait video, the point $(x, y)$ in $I_t$ would become ${(x+\Delta x, y +\Delta y)}$ in ${{I}_{t+\Delta t}}$, and the brightness remains unchanged:
\begin{equation}
\label{eq3}
I(x,y,t)=I(x+\Delta x,y+\Delta y,t+\Delta t)
\end{equation}

Likewise, the definition of optical flow and this kind of assumption can be generalized from the image level to the feature level:
\begin{equation}
\label{eq4}
\frac{\partial {{f}_{sm}}(x,y,t)}{\partial x}{{v}_{x}}+\frac{\partial {{f}_{sm}}(x,y,t)}{\partial y}{{v}_{y}}+\frac{\partial {{f}_{sm}}(x,y,t)}{\partial t}=0
\end{equation}
where ${{f}_{sm}}(x,y,t)$ is the vector at location $(x, y, t)$ in the features $f_{sm}$. The computing to generate $f_{sm}$ is realized by convolutional layers, which are differentiable. Consequently, from Eq. \ref{eq4}, we can get the motion-oriented features:
\begin{equation}
\label{eq5}
{{f}_{mo}}(x,y,t)\!=\!\left[ \frac{\partial\! {{f}_{sm}}(x,\!y,\!t)}{\partial x},\! \right.\frac{\partial\! {{f}_{sm}}(x,\!y,\!t)}{\partial y},\!\left. \frac{\partial\! {{f}_{sm}}(x,\!y,\!t)}{\partial t} \right]
\end{equation}

It can be seen that ${{f}_{mo}}(x,y,t)$ consists of three gradients along $x$, $y$, and $t$ axis independently. This group of features integrates spatial and temporal information, effectively reflecting the motion in the video. Moreover, irrelevant information can be filtered by the network during extracting frame-level features, so the feature-level tensor is less affected by noise than the raw silhouettes. Hereby, the gait pattern can be better encoded.

\begin{figure}[!t]
\centering
\includegraphics[width=7cm]{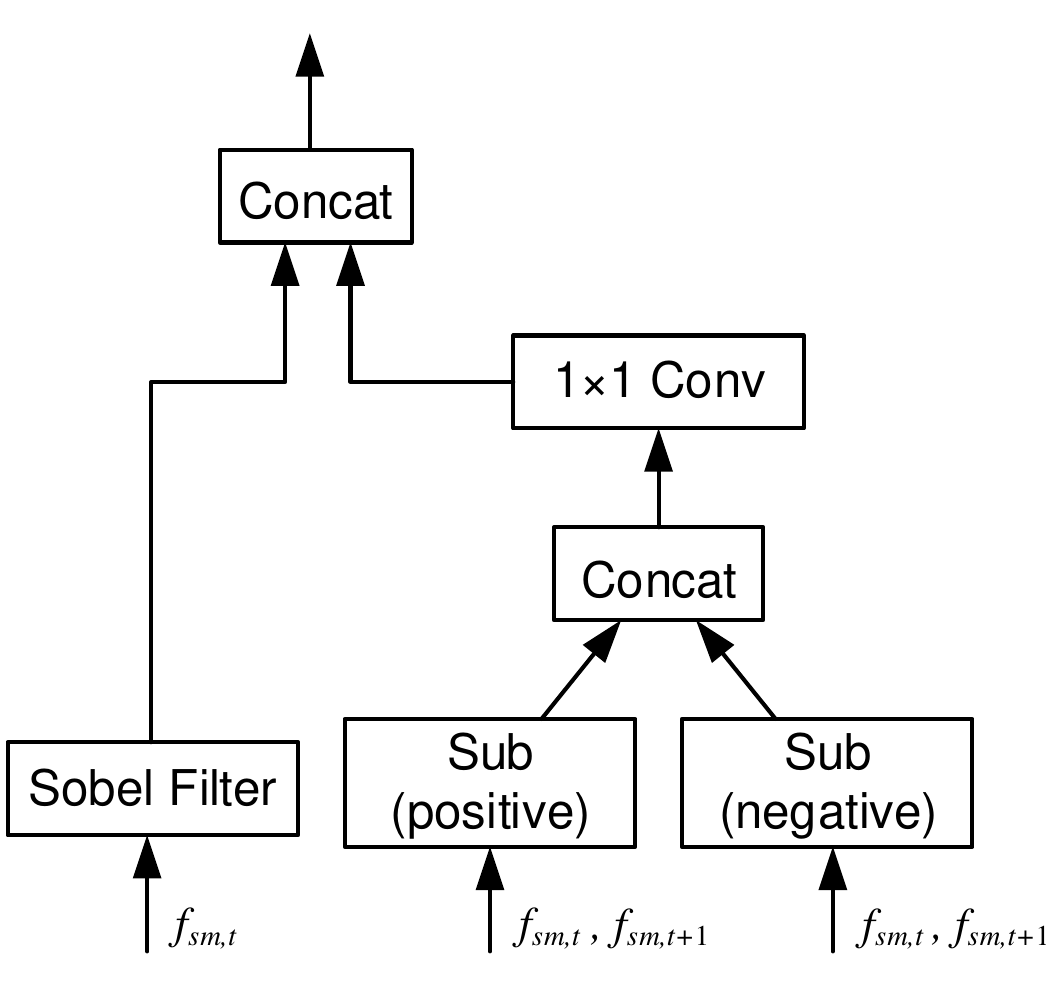}
\caption{Bilateral motion-oriented basic sub-block.}
\label{fig2}
\end{figure}

Three gradients, $G_x$, $G_y$, and $G_t$, are the main ingredients of ${{f}_{mo}}(x,y,t)$. Traditional methods and artificial neural networks are combined to calculate the $f_{mo}$ here, whose structure is shown in Fig. \ref{fig2}. Typically, the time domain derivative can be formulated as:
\begin{equation}
\label{eq6}
\Delta \text{t=}D(x,y,t)\text{=}{{f}_{sm}}(x,y,t+1)-{{f}_{sm}}(x,y,t)
\end{equation}

Apparently, there are negative results after subtraction, and the sign denotes the direction of the motion. To be specific, negative ones indicate where human body parts have moved away, while positive values indicate where they will appear. However, such results are not suitable for CNNs to understand, as CNNs are designed for image processing and the images are totally composed of positive numbers. In this case, CNNs regard the positive and negative signs as a hint of the numerical magnitude, rather than the direction. Additionally, some popular activation functions, like ReLU and leak ReLU, would further undermine the impact of negative results. As a consequence, we separate the difference, getting the positive one and the negative one individually. Then we take the absolute value of both of them, and exploit a convolutional layer with 1×1 kernels to fuse the two parts of ${\Delta t}$ to compute bilateral $G_t$:
\begin{align}
D{{(x,y,t)}_{+}}\text{=D}(x,y,t),\;  p(x,y)>\text{0} \\
D{{(x,y,t)}_{-}}\text{=D}(x,y,t),\;  p(x,y)<\text{0}
\end{align}
\begin{equation}
\label{eq9}
{{G}_{t}}=Conv1(Concat\{D{{(x,y,t)}_{+}},D{{(x,y,t)}_{-}}\})
\end{equation}
where $D{{(x,y,t)}_{+}}$ and $D{{(x,y,t)}_{-}}$ denotes the positive and negative features in the difference, and $p(x,y)$ is the pixel value of point $(x, y)$. In this case, given that the channel dimension of ${{f}_{sm}}(x,y,t)$ is $c$, the input channel dimension of $Conv1(\cdot)$ is $2c$ and the output channel dimension is $c$, consistent with the other two gradients. The right part of Fig. \ref{fig2} explains this detailed procedure. Since $G_t$ is bilateral, we fully take advantage of temporal information.

For $G_x$ and $G_y$, they are spatial representations on $x$ and $y$ axis, and Sobel filters can be utilized to generate them:
\begin{equation}
\label{eq10}
{{G}_{x}}=\left[ \begin{matrix}
   \text{-}1 & 0 & \text{1}  \\
   \text{-}2 & 0 & \text{2}  \\
   \text{-}1 & 0 & \text{1}  \\
\end{matrix} \right]*{{f}_{sm}},\; {{G}_{y}}=\left[ \begin{matrix}
   \text{-}1 & \text{-}2 & \text{-}1  \\
   0 & 0 & 0  \\
   \text{1} & \text{2} & \text{1}  \\
\end{matrix} \right]*{{f}_{sm}}
\end{equation}
where $*$ is convolution calculation deployed on every channel in $f_{sm}$. Because the Sobel operator depends on the first-order derivative, it corresponds to the definition of $G_x$ and $G_y$. In this way, the bilateral motion-oriented features, $\text{ }\!\!\{\!\!\text{ }{{G}_{x}},{{G}_{y}},{{G}_{t}}\text{ }\!\!\}\!\!\text{ }$, can be generated. Finally, the three gradients are concatenated in the channel dimension
\subsection{Temporal multi-scale features}\label{3.3}
Spatial features usually have various resolutions. As a network expands in depth, the features tend to become refined and have a larger receptive field. Exploiting multi-scale features can refine the depiction of the input, and such algorithms have proved their effectiveness in the tasks of classification, detection, and segmentation.

\begin{figure}[!t]
\centering
\includegraphics[width=7.8cm]{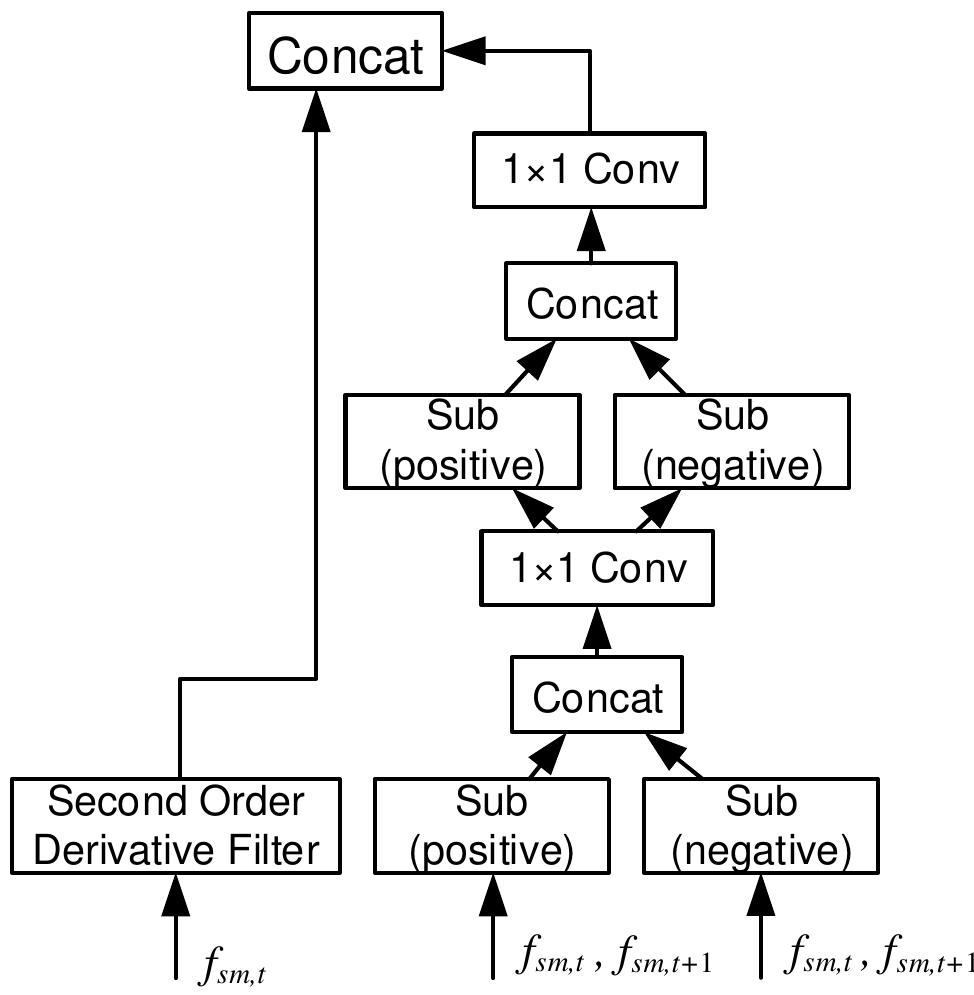}
\caption{Bilateral motion-oriented sub-block for the second scale. (The embedded approach based on second-order derivatives)}
\label{fig3}
\end{figure}

Correspondingly, multi-scale spatiotemporal features would comprehensively convey gait dynamics. ${{f}_{mo}}(x,y,t)$ mentioned in Section \ref{3.2} can be considered as coarse-grained features, since only the first-order motion features are calculated. This also can be explained by the definition of optical flow. Optical flow describes the motion by velocity, which is a representation derived from the first derivative. So ${{f}_{mo}}(x,y,t)$ can be regarded as ${{f}_{mos1}}(x,y,t)$ that comes from the first temporal scale. From this perspective, a second-order descriptor can be viewed as a type of acceleration tensor to depict the motion, whereas a high-order descriptor can be seen as a kind of more refined temporal representation.

Moreover, a deeper network with several bilateral motion-oriented basic sub-blocks can provide refined features at a high temporal level. Thus, features on the second scale can be generated through two stacked basic sub-blocks shown in Fig. \ref{fig2}. As ${{f}_{mo}}(x,y,t)={{f}_{mos1}}(x,y,t)=M({{f}_{sm,t}},{{f}_{sm,t+1}})$, ${{f}_{mos2}}(x,y,t)$ can be easily acquired as:

\begin{equation}
\begin{aligned}
\label{eq11}
{{f}_{mos2}}(x,y,t)& =M(Conv1({{f}_{mos1}}(x,y,t)) \\ 
& =M(Conv1(M({{f}_{sm,t}},{{f}_{sm,t+1}})))  
\end{aligned}   
\end{equation}
where $M(\cdot)$ is the function of operation to obtain $\text{ }\!\!\{\!\!\text{ }{{G}_{x}},{{G}_{y}},{{G}_{t}}\text{ }\!\!\}\!\!\text{ }$ (Eq. \ref{eq9} and Eq. \ref{eq10}), whose process is illustrated in Fig. \ref{fig2}, and $Conv1(\cdot)$ is a convolutional layer with 1×1 kernels. The goal of $Conv1(\cdot)$ is to shrink the channel dimension of ${{f}_{mos1}}(x,y,t)$ to reduce the computational burden, because $M(\cdot)$ triples the dimension of the input. 

However, this process containing many convolutional layers might be hard to train, so we propose another scheme here, a simple embedded approach based on high-order gradients. Similar to Eq. \ref{eq5}, the second-order features ${{f}_{mos2}}(x,y,t)$ can be formulated as:

\begin{equation}
\label{eq12}
{{f}_{mos2}}(x,\! y,\! t)\!=\!\left[\! \frac{{{\partial }^{2}}\! {{f}_{sm}}(x,\! y,\! t)}{\partial {{x}^{2}}},\!  \right.\frac{{{\partial }^{2}}\!{{f}_{sm}}(x,\! y,\! t)}{\partial {{y}^{2}}},\! \left. \frac{{{\partial }^{2}}{{f}_{sm}}(x,y,t)}{\partial {{t}^{2}}} \! \right]
\end{equation}

Likewise, thus three second-order gradients, ${{G}_{xx}}$, ${{G}_{yy}}$, and ${{G}_{tt}}$, are the main components of ${{f}_{mos2}}(x,y,t)$. Fig. \ref{fig3} explains the specific implementation of the embedded operation, where two bilateral difference units are served to represent ${{G}_{tt}}$, while spatial features, ${{G}_{xx}}$ and ${{G}_{yy}}$, can be calculated by the second-order derivative:
\begin{equation}
\label{eq13}
{{G}_{xx}}\!=\!{{f}_{sm,t}}(x\!-\!1,y,t)\!-\!2{{f}_{sm,t}}(x,y,t)\!+\!{{f}_{sm,t}}(x\!+\!1,y,t)
\end{equation}
\begin{equation}
\label{eq14}
{{G}_{yy}}\!=\!{{f}_{sm,t}}(x,y\!-\!1,t)\!-\!2{{f}_{sm,t}}(x,y,t)\!+\!{{f}_{sm,t}}(x,y\!+\!1,t)
\end{equation}

We extend Eq. \ref{eq13} and Eq. \ref{eq14} to the convolutional calculation with 3×3 kernels, then have:
\begin{equation}
\label{eq15}
{{G}_{xx}}=\left[ \begin{matrix}
   1 & \text{-}2 & 1  \\
   2 & \text{-4} & 2  \\
   1 & \text{-}2 & 1  \\
\end{matrix} \right]*{{f}_{sm}},\;  {{G}_{yy}}=\left[ \begin{matrix}
   1 & 2 & 1  \\
   \text{-}2 & \text{-4} & \text{-}2  \\
   1 & 2 & 1  \\
\end{matrix} \right]*{{f}_{sm}}
\end{equation}

Similar to Eq. \ref{eq10}, this operation is deployed on every channel in $f_{sm}$. Hereby, the second-order spatial derivative filters can be implemented by Eq. \ref{eq15}. In conclusion, we put forward two kinds of ways to generate ${{f}_{mos2}}(x,y,t)$, a stacking approach and an embedded approach, formulated by Eq. \ref{eq11} and Eq. \ref{eq12} independently.

Higher-level temporal features can be achieved by combining two stacked sub-blocks shown in Fig. \ref{fig2} and Fig. \ref{fig3}. For example, stacking the basic sub-block and the second scale sub-block would generate the features at the third scale, and two stacked sub-blocks shown in Fig. \ref{fig3} would generate the features at the fourth scale. By this means, the motion-oriented features can be generalized as a set ${{\tilde{f}}_{mo,t}}=\{{{f}_{mos1,t}},{{f}_{mos2,t}},\ldots ,{{f}_{mosi,t}}\}$.
\subsection{Spatial correction block}
Video-based gait recognition is always built on a special data type, silhouettes, since the data after foreground and background segmentation contribute to eliminating the distraction of light and background, enforcing the algorithm to concentrate on humans. Silhouettes, however, are sometimes imprecise, owing to the low resolution of the video as a result of the long distance between the camera and the subject. Many appearance-based gait recognition researches solve this problem by consolidating information from the entire video. For instance, the global max temporal pooling can find out the frame with accurate segmentation\cite{RN1,RN15}. Nevertheless, motion-based methods would amplify the segmentation noise, as both segmentation noise and motion are reflected at the outer edge of silhouettes. In this case, this segmentation noise is treated as motion dynamics, which would confuse the network.  

\begin{figure}[!t]
\centering
\includegraphics[width=8.5cm]{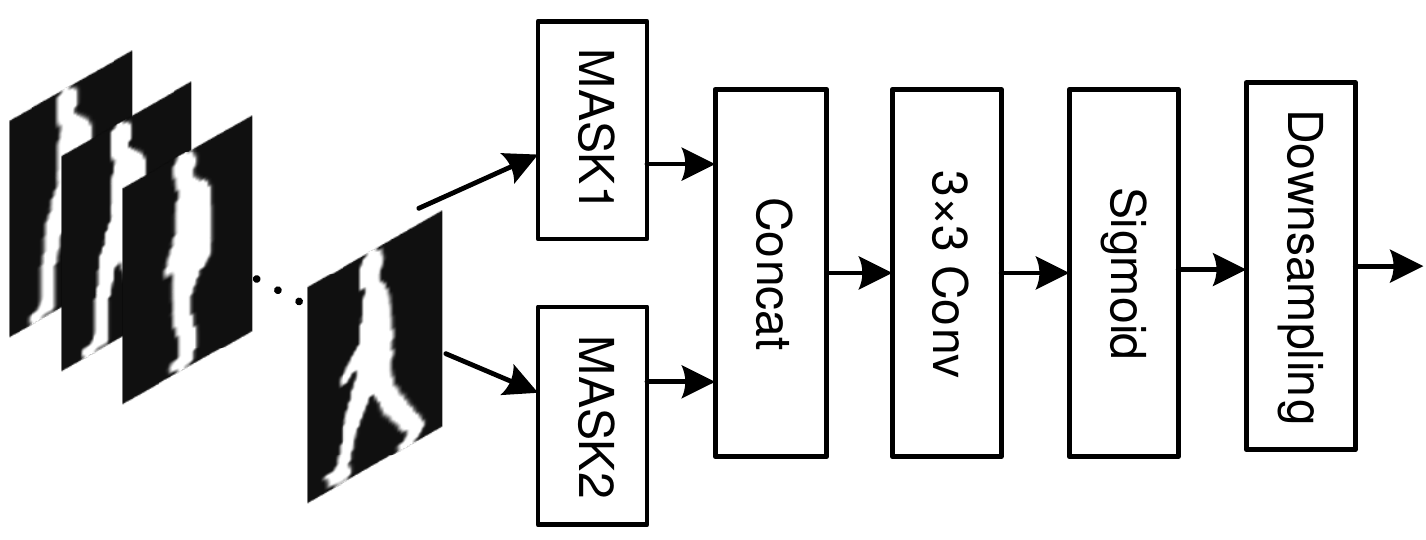}
\caption{Spatial correction block.}
\label{fig4}
\end{figure}

\begin{figure}[!t]
\centering
\includegraphics[width=8.8cm]{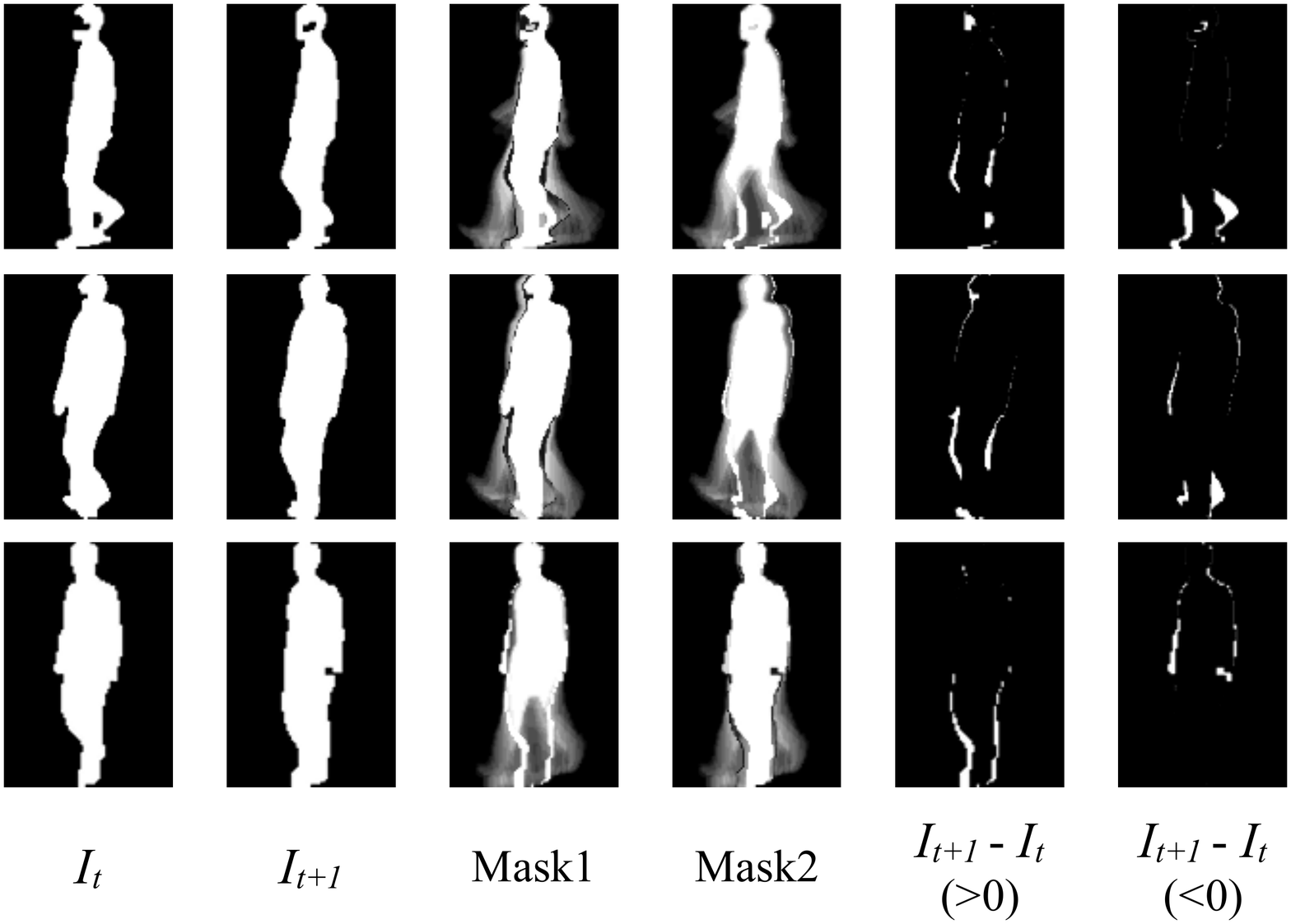}
\caption{Samples of the masks.}
\label{fig5}
\end{figure}

To alleviate this limitation, we propose a spatial correction block to evaluate the accuracy of segmentation as shown in Fig. \ref{fig4}, where the whole video information is adopted. In particular, we construct two masks from the raw frames, $mask1=\left| {{I}_{t}}+{{I}_{t+1}}-mean(\tilde{I}) \right|$ and $mask2=\left| {{I}_{t+1}}-{{I}_{t}} \right|+mean(\tilde{I})$. Since there is just one subject in a video, all frames have the same static (appearance) information. So the universal and common parts $mean(\tilde{I})$ can be considered as the static information, while $\left| {{I}_{t+1}}-{{I}_{t}} \right|$ can be considered as the dynamic information. Thereby, $mask2$ has static and dynamic features. For $mask1$, there is the dynamic part in ${{I}_{t}}+{{I}_{t+1}}$, but the static also exists in both $I_t$ and $I_{t+1}$, so $mean(\tilde{I})$ is subtracted. Intuitively, by measuring the similarity of the two masks, the accuracy of the segmentation can be assessed. Fig. \ref{fig5} exhibits some samples of masks to visualize their functionalities. It can be found that the big difference 
at the edges between the two frames indicates the motion as shown in the last two columns. For the samples in Fig. \ref{fig5}, there is the obvious inexact segmentation in $I_t$ and $I_{t+1}$, and $mask1$ and $mask2$ have significant differences at the location where the segmentation is incorrect. Therefore, less similarity would give smaller weights to the motion features of this part, thereby correcting the segmentation error.

As shown in Fig. \ref{fig4}, metrics and weight assignment are carried out by a convolutional layer and a sigmoid function. The masks are concatenated on the channel axis, and the number of channels becomes one after the convolutional layer, then the weight is yielded through the sigmoid. Finally, the output is down-sampled to the same size of ${{\tilde{f}}_{mo,t}}$. Next, the weights are assigned to ${{\tilde{f}}_{mo,t}}$ by a point-by-point multiplication as shown in Fig. \ref{fig1}.

\section{Experiments}
The performance of the proposed method has been evaluated on the CASIA-B dataset\cite{RN14} and the OU-MVLP dataset\cite{RN13}. First, SMBM is compared with the recent gait recognition methods on CASIA-B and OU-MVLP. Then, a comprehensive ablation study is reported on CASIA-B to demonstrate the effectiveness of each component (bilateral motion-oriented features, temporal multi-scale features, and spatial correction block). 
\subsection{Dataset and protocol}
The CASIA-B dataset is one of the widely used gait datasets composed of 124 participants, each comprising 11 walking observation points: 0°, 18°, 36°, ..., 180°, as well as three walking conditions: normal condition, carrying bags, and changing clothes\cite{RN14}. Under each view, every person has 10 sequences on any given view (four sequences for the normal condition, two for carrying bags, and two for altering clothes (wearing jackets)). In total, there are 110 sequences (10 × 11 = 110) of a subject in this dataset. In this study, a commonly adopted test protocol is applied, where the first 74 subjects are designated as the training set, and the remaining 50 subjects are reserved for testing\cite{RN1,RN56,RN58}. During testing, the first four sequences of NM (NM\#1-4) are selected as the gallery set, while the remaining six sequences are served as the test set. Moreover, the test set is divided into three subsets: the NM subset (consisting of NM\#5-6), the BG subset (consisting of BG 1-2), and the CL subset (consisting of CL\#1-2).

The OU-MVLP dataset is one of the largest public gait datasets, including 10307 subjects\cite{RN13}. There are 14 different walking directions, from 0° to 90° (0°,15°,...,90°), and 180° to 270° (180°,195°,...,270°), and two sequences (\#00 and \#01) for each view. We here exploit a prevalent testing scheme, in which four views, 0°, 30°, 60°, and 90°, are deployed for cross-view gait recognition\cite{RN56,RN10,RN25}. In this protocol, 5153 subjects are placed as the training set and the remaining 5154 subjects are reserved for testing. During testing, the first sequence (\#00) of each subject constitutes the gallery set, while the rest of the sequences (\#01) is served as the test set.

\subsection{Implementation details}

\subsubsection{Loss}
We jointly employ the triplet loss and the focal loss to drive the model to achieve the balance between easy and difficult samples under the complicated scenarios. Incorporating these two losses helps the network to devote more attention to challenging samples.

Batch all triplet loss is strongly suitable for gait recognition, which aims to measure the distance in positive and negative samples\cite{RN15,RN19}.
\begin{equation}
\label{eq16}
{{L}_{triplet}}=\frac{Max(M-L{{2}_{anch,pos}}+L{{2}_{anch,neg}},0)}{2M}
\end{equation}
where $M$ is a defined margin, which intends to minimize the distance between the features belonging to the same subject during training. To enhance the learning capability, the cross-entropy loss is usually included to aid in the discovery of a more discriminant gait metric space\cite{RN65}. However, there are numerous difficult samples in gait recognition, such as samples from different views (clothes or bags). Enhancing the learning of these samples is able to strengthen the robustness of the network, so we exploit the focal loss\cite{RN87}. It can decrease the weights of easy samples, thereby focusing on hard ones in the training phase:
\begin{equation}
\label{eq17}
{{L}_{focal}}=-{{(1-p)}^{\gamma }}\log (p)
\end{equation}
where $p$ is the estimated probability of each class, obtained by the softmax function, and ${\gamma}$ is a defined parameter to adjust the rate smoothly. When $p$ is small, i.e., the video is misclassified, ${{{(1-p)}^{\gamma}}}$ approaches 1 and the loss approaches the CE loss. Conversely, when $p$ is 1, ${{{(1-p)}^{\gamma}}}$ becomes 0 and the loss of well-classified samples is decreased. Therefore, the total loss is formulated as:
\begin{equation}
\label{eq18}
L={{L}_{triplet}}+\lambda {{L}_{focal}}
\end{equation}
where ${\lambda}$ is a parameter to balance the two losses.

\subsubsection{Network structure}
As shown in Fig. \ref{fig1}, there are two branches in the SMBM network. The structure of the spatiotemporal multi-scale bilateral motion branch has been explained in the Section \ref{3}, and its parameters will be examined in the ablation experiments (Section \ref{Ablation experiments}). The backbone of another branch is composed of three CNN blocks, whose details are illustrated in Table~\ref{table1}. Conv\_64\_5 denotes a convolutional layer with 5×5 kernels, and its output channel number is 64. By the same way, the setting of other convolutional layers in Table~\ref{table1} can be clearly defined and understood. LReLU means the Leaky ReLU activation function, and MaxPool\_2 is the Max pooling with 2×2 kernels.

In addition, we apply global max pooling for temporal pooling. In HPM, the feature map is divided into 1, 2, 4, 8, and 16 strips respectively\cite{RN1}, HPMs individually enable the strip to map as a vector of length 256 for the static features and the motion-oriented features. The last layers shown in Fig. \ref{fig1}, FC with normalization, contains the neurons equal to the number of classes in the training stage. Therefore, it should be 74 for CASIA-B and 5153 for OU-MVLP.
\begin{table}[!t]
\caption{CNN Block Structure\label{table1}}
\centering
\begin{tabular}{|c|c|}
\hline
Block                             & Layers              \\
\hline\multirow{3}{*}{CNN Block1} & Conv\_64\_5         \\
                                  & Conv\_64\_3\_LReLU  \\
                                  & MaxPool\_2          \\
\hline\multirow{3}{*}{CNN Block2} & Conv\_128\_3        \\
                                  & Conv\_128\_3\_LReLU \\
                                  & MaxPool\_2          \\
\hline\multirow{2}{*}{CNN Block3} & Conv\_256\_3        \\
                                  & Conv\_256\_3\_LReLU \\
\hline
\end{tabular}
\end{table}

\subsubsection{Training and test details}
First, the gait bounding box of the input contour is aligned and resized to 64 × 44. The length of the video in the training phase is set to 30. In another word, 30 ordered frames in a sequence are selected randomly. Specifically, original sequences with fewer than 15 frames should be deleted, whereas sequences with more than 15 but fewer than 30 frames should be repeatedly sampled. $M$  is placed as 0.2 in the triplet loss. The optimizer is Adam with a momentum of 0.9. Batch all triplet loss is exploited in training\cite{RN19}, so the batch size is $P\times{K}$, $P$ indicates the number of identities and $K$ denotes the number of samples per class in a batch. ${\lambda}$, batch size, learning rate, and the number of iterations are different for the two datasets, due to the difference in data volume of them. For CASIA-B, ${\lambda}$ is 1, and the batch size is set as (8,8) constrained by hardware. The number of iterations is 60K. The learning rate starts at 0.1,  decreases to 0.01 after the 20thK iterations, and then drops to 0.001 after the 40thK iterations, eventually sets to 1e-4 after the 50K iterations. For OU-MVLP, ${\lambda}$ is 0.1, since there are a much greater number of classes that make the last FC layer extremely complicated. A larger ${\lambda}$ leads to challenging training. Through the experiment, the loss does not converge, and the network cannot be trained when ${\lambda}$ is bigger than 0.1. The batch size is (16,4) because of the hardware limitation. The number of iterations is 550K. The learning rate is firstly 0.1, and drops to 0.01 after the 150Kth iteration, then reduces to 0.001 after the 300thK iterations, finally declines to 1e-4 after the 450Kth iteration.

During the testing phase, gait sequences can be fed directly into the model, and the videos including fewer than five frames are discarded. This is because the temporal refined features require five frames to compute at least. Then the Euclidean distance of the corresponding feature maps between the gallery and the probe is measured. Lastly, we can obtain the identity of the probe subject.
\subsection{Comparison with other methods}
\subsubsection{Evaluation on CASIA-B}
\begin{table*}[!t]
\caption{Comparison with Other Methods under NM on CASIA-B, excluding identical-view cases.\label{table2}}
\centering
\begin{tabular}{|c|c|c|c|c|c|c|c|c|c|c|c|c|c|c|c|c|}
\hline
\multirow{2}{*}{Methods}      & \multicolumn{11}{c|}{Probe   View (\%)}                                                                                                                                       & \multirow{2}{*}{Mean} \\
\cline{2-12}& 0°            & 18°           & 36°            & 54°           & 72°           & 90°           & 108°          & 126°          & 144°          & 162°          & 180°        &                       \\
\hline
\cite{RN2}  & 82.6          & 90.3          & 96.1           & 94.3          & 90.1          & 87.4          & 89.9          & 94.0          & 94.7          & 91.3          & 78.5        & 89.9                  \\
\cite{RN21} & 75.6          & 91.3          & 91.2           & 92.9          & 92.5          & 91.0          & 91.8          & 93.8          & 92.9          & 94.1          & 81.9        & 89.9                  \\
\cite{RN1}  & 90.8          & 97.9          & 99.4           & 96.9          & 93.6          & 91.7          & 95.0          & 97.8          & 98.9          & 96.8          & 85.8        & 95.0                  \\
\cite{RN8}  & 93.1          & 92.6          & 90.8           & 92.4          & 87.6          & {\ul 95.1}    & 94.2          & 95.8          & 92.6          & 90.4          & 90.2        & 92.3                  \\
\cite{RN71} & 92.0          & 98.5          & \textbf{100.0} & {\ul98.9} & 95.7          & 91.5          & 94.5          & 97.7          & 98.4          & 96.7          & 91.9        & 96.0                  \\
\cite{RN10} & 91.1          & 98.0          & 99.4           & 98.2          & 93.2          & 91.9          & 95.2          & 98.3          & 98.4          & 95.7          & 87.5        & 95.2                  \\
\cite{RN40} & 55.3          & 93.6          & 73.9           & 75.0          & 68.0          & 68.2          & 71.1          & 72.9          & 76.1          & 70.4          & 55.4        & 68.7                  \\
\cite{RN67} & 92.3          & 93.2          & 92.9           & 93.9          & 91.9          & 94.1          & 94.3          & 93.3          & 92.8          & 91.1          & 91.1        & 92.8                  \\
\cite{RN15} & 94.1          & 98.6          & 99.3           & 98.5          & 94.0          & 92.3          & 95.9          & 98.4          & 99.2          & 97.8          & 90.4        & 96.2                  \\
\cite{RN56} & 89.7          & 98.5          & {\ul 99.8}     & 97.9          & 94.4          & 91.2          & 94.5          & 97.1          & 97.6          & 97.0          & 89.4        & 95.2                  \\
\cite{RN57} & \textbf{96.0} & 96.2          & 97.3           & 96.1          & 93.9          & 91.8          & 93.0          & 95.7          & 96.6          & 97.3          & \textbf{94.0} & 95.3                  \\
\cite{RN58} & 91.8          & 98.3          & 99.0           & 98.0          & 94.1          & 92.8          & 96.3          & 98.1          & 98.4          & 96.2          & 89.2        & 95.7                  \\
\cite{RN52} & 93.2          & 99.3 & 99.5           & 98.7          & \textbf{96.1} & \textbf{95.6} & \textbf{97.2} & 98.1          &  99.3    & \textbf{98.6} & 90.1        & {\ul 96.9}            \\
\cite{RN59} & {\ul 95.7}    & 98.2          & 99.0           & 97.5          & 95.1          & 93.9          & 96.1          & {\ul 98.6}    & 99.2          & 98.2          & 92          & 96.7                  \\
\cite{RN70} & 95.1          & {\u99.0}    & 99.1           & 98.3          & 95.7          & 93.6          & 95.9          & 98.3          & 98.6          & 97.7          & 90.8        & 96.6                  \\
\cite{RN16} & 95.5 &\textbf{99.2} &99.6 &\textbf{99.0} &94.4 &92.5 &95.0 &98.1 &{\ul99.7} &98.3 &{\ul92.9} &96.7                  \\
Ours                         & 94.5          &  {\ul99.0}    & 99.6           & {\ul98.9} & \textbf{96.1} & 93.2          & {\ul 97.1}    & \textbf{98.8} & \textbf{99.8} & {\ul 98.5}    & {\ul 92.9}  & \textbf{97.1}        \\
\hline
\end{tabular}
\end{table*}

\begin{table*}[!t]
\caption{Comparison with Other Methods under BG on CASIA-B, excluding identical-view cases.\label{table3}}
\centering
\begin{tabular}{|c|c|c|c|c|c|c|c|c|c|c|c|c|c|c|c|c|}
\hline
\multirow{2}{*}{Methods}      & \multicolumn{11}{c|}{Probe   View (\%)}                                                                                                                                       & \multirow{2}{*}{Mean} \\
\cline{2-12}& 0°            & 18°           & 36°            & 54°           & 72°           & 90°           & 108°          & 126°          & 144°          & 162°          & 180°        &                       \\
\hline
\cite{RN2}  & 64.2          & 80.6          & 82.7          & 76.9          & 64.8          & 63.1          & 68.0          & 76.9          & 82.2          & 75.4          & 61.3          & 72.4          \\
\cite{RN1}  & 83.8          & 91.2          & 91.8          & 88.8          & 83.3          & 81.0          & 84.1          & 90.0          & 92.2          & 94.4          & 79.0          & 87.2          \\
\cite{RN8}  & 88.8          & 88.7          & 88.7          & 94.3          & 85.4          & 92.7          & 91.1          & 92.6          & 84.9          & 84.4          &  86.7    & 88.9          \\
\cite{RN10} & 86.0          & 93.3          & 95.1          & 92.1          & 88.0          & 82.3          & 87.0          & 94.2          & 95.9          & 90.7          & 82.4          & 89.7          \\
\cite{RN67} & 87.3          & 85.5          & 85.0          & 84.1          & 82.3          & 82.9          & 84.6          & 82.7          & 81.7          & 85.6          & 82.4          & 84.0          \\
\cite{RN15} & 89.1          & 94.8          & {\ul 96.7}    & \textbf{95.1} & 88.3          & 84.9          & 89.0          & 93.5          & 95.1          & 93.8          & 85.8          & 91.5          \\
\cite{RN56} & 86.7          & 94.6          & 96.0          & 92.5          & 85.8          & 80.5          & 84.9          & 91.5          & 96.0          & 93.1          & 86.0          & 89.8          \\
\cite{RN58} & 87.3          & 93.7          & 94.8          & 93.1          & 88.1          & 84.5          & 88.8          & 93.5          & 96.3          & 93.3          & 83.9          & 90.7          \\
\cite{RN59} & 91.0          & 95.4          & \textbf{97.5} & 94.2          & \textbf{92.3} & 86.9          & \textbf{91.2} & \textbf{95.6} & \textbf{97.3} & {\ul 96.4}    & 86.6          & {\ul 93.0}      \\
\cite{RN70} & {\ul 92.3}    & \textbf{96.6} & 96.6          & 94.5          & {\ul 91.9}    & \textbf{87.6} & 90.7          & {\ul 94.7}    & 96.0          & 93.9          & 86.1          & 92.8          \\
\cite{RN16} & 90.2 &{\ul96.4} &96.1 &{\ul94.9} &89.3 &85.0 &90.9 &94.5 &96.3 &95.0 &\textbf{88.1} &92.4          \\
Ours                         & \textbf{93.1} &  95.8    & 96.0            & 94.6    & 91.7          & {\ul 87.0}      & {\ul 91.1}    & 94.3          & {\ul 97.2}    & \textbf{96.5} & {\ul87.4} & \textbf{93.2}\\
\hline
\end{tabular}
\end{table*}

\begin{table*}[!t]
\caption{Comparison with Other Methods under CL on CASIA-B, excluding identical-view cases.\label{table4}}
\centering
\begin{tabular}{|c|c|c|c|c|c|c|c|c|c|c|c|c|c|c|c|c|}
\hline
\multirow{2}{*}{Methods}      & \multicolumn{11}{c|}{Probe   View (\%)}                                                                                                                                       & \multirow{2}{*}{Mean} \\
\cline{2-12}& 0°            & 18°           & 36°            & 54°           & 72°           & 90°           & 108°          & 126°          & 144°          & 162°          & 180°        &                       \\
\hline
\cite{RN2}  & 37.7        & 57.2          & 66.6          & 61.1        & 55.2          & 54.6          & 55.2        & 59.1          & 58.9          & 48.8          & 39.4          & 54.0          \\
\cite{RN1}  & 61.4        & 75.4          & 80.7          & 77.3        & 72.0          & 70.1          & 71.5        & 73.5          & 73.5          & 68.4          & 50.0          & 70.4          \\
\cite{RN8}  & 42.1        & 58.2          & 65.1          & 70.7        & 68.0          & 70.6          & 65.3        & 69.4          & 51.5          & 50.1          & 36.6          & 58.9          \\
\cite{RN10} & 65.8        & 80.7          & 82.5          & 81.1        & 72.7          & 71.5          & 74.3        & 74.6          & 78.7          & 75.8          & 64.4          & 74.7          \\
\cite{RN67} & 50.1        & 60.7          & 72.4          & 72.1        & 74.6          & \textbf{78.4} & 70.3        & 68.2          & 53.5          & 44.1          & 40.8          & 62.3          \\
\cite{RN15} & 70.7        & 85.5          & 86.9          & 83.3        & 77.1          & 72.5          & 76.9        & 82.2          & 83.8          & 80.2          & 66.5          & 78.7          \\
\cite{RN56} & 63.7        & 79.2          & 82.3          & 77.7        & 69.4          & 71.5          & 73.5        & 77.9          & 78.4          & 76.5          & 62.4          & 73.9          \\
\cite{RN58} & 63.4        & 77.3          & 80.1          & 79.4        & 72.4          & 69.8          & 71.2        & 73.8          & 75.5          & 71.7          & 62.0          & 72.4          \\
\cite{RN59} & \textbf{76.0} & \textbf{87.6} & {\ul89.8} & {\ul85.0} & \textbf{81.2} & 75.7          & \textbf{81.0} & \textbf{84.5} & \textbf{85.4} &  82.2    &  68.1    & {\ul81.5} \\
\cite{RN70} & {\ul 75.6}  & 87.1          &  88.3    & 83.1        & 78.8          & {\ul 78.0}      & {\ul 79.9}  & 82.7          &  83.9    & 78.9          & 66.6          & 80.3          \\
\cite{RN16} & 75.6 &87.0 &\textbf{88.9} &\textbf{86.5} &{\ul 80.5} &77.5 &79.1 &{\ul 84.0} &{\ul84.8} &{\ul83.6} &\textbf{70.1} &\textbf{81.6}          \\
Ours                         & 71.9        & {\ul 87.3}    & 87.6          &  83.8  & 79.5    & 77.2          & 78.5        & 82.8    & 83.8          & \textbf{84.2} & {\ul68.8} &  80.5   \\
\hline
\end{tabular}
\end{table*}

Table~\ref{table2} reports the averaged view-crossed rank-1 accuracies compared with other current researches under the normal condition. The probe samples are tested on all 11 views, and the gallery samples come from all views except the corresponding identical view. The results shown in Table~\ref{table2} are directly derived from their original papers. For a clear presentation, the best and suboptimal results in the table are separately bolded and underlined for highlighting.

From Table~\ref{table2}, it is clear that: (1) The proposed network yielded the optimal mean accuracy, 97.1\%, which is the only approach that exceeds 97\% so far. (2) On the probe view of 54°, 72°, 126°, and 144°, our method surpasses others, achieving a remarkable result, nearly 99\%. This might be because middle views (oblique views between the front and the side) can better present the silhouettes changing that can represent the motion. From the front and the rearview, the movement in the forward direction, the key motion during walking, is difficult to be observed. While from the side view, the movement is tricky to distinguish between the left part and the right part. \cite{RN57} and \cite{RN52} show excellent results in other views. They give more attention to the features of special form, respectively taking advantage of information from RGB images or silhouette-level and set-level feature pyramids. (3) \cite{RN2} applies GEI to integrate the video, and some sequential information, especially motion information, would be lost during this process, ending up a relatively weak performance. (4) \cite{RN21}, \cite{RN1}, \cite{RN71}, \cite{RN10}, \cite{RN58}, \cite{RN52}, and \cite{RN70} treat the gait video as a set of images. Although these approaches design the specialized network to analyze and understand human appearance, they essentially neglect the connection between frames, which suggests gait habits. Consequently, this constraint might limit its ability to reach higher performance. (5) \cite{RN40}, \cite{RN67}, and \cite{RN57} employ the skeleton to acquire the abstract motion dynamics effectively, but they might overly rely on the precision of skeleton extraction studies, and the wrong imperceptible skeleton results give rise to misguided learning. Besides, the added RGB input for skeleton extraction complicates the method. (6) LSTM are deployed to capture the gait motion in \cite{RN8}, but this network is complex and difficult to train, and it might focus on the end of the video owing to forgetting. This drawback would restrict the improvement of accuracy. (7) \cite{RN15} and \cite{RN16} assign weights to all frames according to their importance, so it contributes to integrating information in the time domain. However, since there is also inherently no analysis of the connection between frames, the motion is not taken into account. (8) \cite{RN56} represents the motion by the difference, which is simple and efficient, but this approach is susceptible to the segmentation noise and only distills the very local temporal information. (9) \cite{RN59} exploits the 3D-CNN to learn the gait habit. It is a data-driving means to teach the network to comprehend the motion modality. In contrast, our method comes from another perspective, proposing the motion-oriented description of gait information, with solid theoretical interpretability.

Table~\ref{table3} and Table~\ref{table4} compare the performance under complex scenarios, carrying bags and changing clothes. It can be noticed that the proposed method also produces the best results when carrying bags, and the average accuracy reaches 93.2\%, as shown in Table~\ref{table3}. This method achieves outstanding performance at 0°, and 162°, and the possible reason is that silhouettes are less disturbed by bags under such views. Moreover, \cite{RN10}, \cite{RN15}, \cite{RN70}, and \cite{RN16} have the exceptional achievement, too. Because partition method is robust to the covariates like bags and clothes, as it can combine information from each part. Additionally, it is worth noting that this appearance-based partition approach addresses the problem from a different direction, and it is not contradictory to the method in this work. By fusing partition, our method would effectively deal with gait recognition in complicated scenarios in the future. As shown in Table~\ref{table4}, \cite{RN59}, \cite{RN70},  \cite{RN59}, and SMBM yield a mean accuracy of over 80\%, which is an impressive result. Especially for \cite{RN59} and \cite{RN16}, adjacent body parts are segmented and associated to limit the effect of clothes in \cite{RN16}. The 3D-CNN extracts the features invariant to external variables in \cite{RN59}, while the motion-oriented features is a 2D structure depending on the static features, which might be more disrupted by clothing. However, the 1\% gap in performance of the proposed method in the these two cases is acceptable. Therefore, our method is competitive with a solid theoretical foundation and good interpretability.

\subsubsection{Evaluation on OU-MVLP}

\begin{table}[!t]
\caption{Comparison with Other Methods on OU-MVLP, excluding identical-view cases.\label{table5}}
\centering
\begin{tabular}{|c|c|c|c|c|c|c|c|}
\hline
\multirow{2}{*}{Methods}      & \multicolumn{4}{c|}{Probe   View} & \multirow{2}{*}{Mean} \\\cline{2-5}
                            & 0°     & 30°    & 60°    & 90°   &                       \\
\hline
\cite{RN2}  & 6.2    & 22.2   & 26.9   & 21.2  & 19.1                  \\
\cite{RN11} & 8.2    & 32.3   & 33.6   & 28.5  & 25.7                  \\
\cite{RN38} & 12.3   & 29.3   & 30.5   & 18.1  & 22.5                  \\
\cite{RN25} & 51.5   & 70.8   & 66.7   & 63.6  & 63.1                  \\
\cite{RN1}  & 77.7   & 86.9   & 85.3   & 83.5  & 83.4                  \\
\cite{RN56} & \textbf{78.6}   & 87.4   & \textbf{85.9}   & 83.2  & 83.8                  \\
\cite{RN58} & 78.3   & \textbf{88.8}   & 85.7   & 85.1  & \textbf{84.5}                  \\
\cite{RN10} & {\ul 78.5}   & {\ul 87.5}   & {\ul85.8}   & {\ul 85.4}  & {\ul 84.3}                  \\
\cite{RN12} & 56.2   &  73.7   & 81.4   &  82.0  &  73.3                  \\
{}Ours      & 78.3   & 87.2   & {\ul 85.8}   & \textbf{85.8}  & {\ul 84.3}                \\
\hline
\end{tabular}
\end{table}

OU-MVLP contains more than 10000 subjects, which are far more than CASIA-B. Hence, we conduct the experiment on this dataset to evaluate the proposed network. (1) As shown in Table~\ref{table5}, our method achieves 84.3\% on the average accuracy. \cite{RN58} and \cite{RN10} present the performance close to ours, 84.5\% and 84.3\% separately. They employ RNNs or gated recurrent units to analyze the correlations among different human parts, while SMBM focuses on learning the motion pattern of gait. Undoubtedly, appearance-based discriminative features could be further added based on the motion-oriented features, so in this way, the performance can be further strengthened. (2) \cite{RN12} also focuses on the dynamical features without part-based modules, so the result does not exceed \cite{RN58} and \cite{RN10}. Compared with our method, an autoencoder is deployed to approximate the Koopman operators in \cite{RN12}, while we design a block to directly extract the optical-flow-like features here. The motion features are intuitively calculated in our paper, and better accuracies are achieved. (3) In addition, \cite{RN11}, \cite{RN1}, \cite{RN56}, \cite{RN58}, and \cite{RN10} deepen and widen the backbone network to accommodate more data in OU-MVLP, but the network architecture is consistent with that deployed for CASIA-B in this study, except for the last layer for classification. This indicates that the SMBM network has great capability of generalization. 

\subsection{Ablation experiments}\label{Ablation experiments}

\begin{table*}[!t]
\caption{Ablation experiments conducted on CASIA-B in terms of various settings on spatiotemporal multi-scale bilateral motion context branch.\label{table6}}
\centering
\begin{tabular}{|c|c|c|c|c|c|c|c|c|c|c|c|c|c|}
\hline
\multirow{2}{*}{Number} & \multicolumn{3}{c|}{Motion features} & \multirow{2}{*}{Multi-scale temporal} & \multirow{2}{*}{Correction} & \multicolumn{4}{c|}{RESULTS}                                   \\
\cline{2-4}   \cline{7-10}
                        & NONE        & MOF         & BMOF   &                                       &                             & NM            & BG            & CL            & Mean          \\
\hline
1                       & \checkmark   &             &        & -                                     & -                           & 96.3          & 92.0          & 77.7          & 88.7          \\
2                       &             & \checkmark   &        & -                                     & -                           & 97.1          & 93.0          & 79.2          & 89.8          \\
3                       &             &             & \checkmark      & -                                     & -                           & \textbf{97.3} & 93.1          & 79.5          & {\ul 90.0}      \\
4                       &             & \checkmark   &        & -                                     & \checkmark   & 97.1          & 92.8          & 79.0          & 89.6          \\
5                       &             &             & \checkmark      & -                                     & \checkmark   & 97.1          & \textbf{93.3} & 79.7          & {\ul 90.0}      \\
6                       &             &             & \checkmark      & 1,2(s)                                & \checkmark   & 96.9          & 92.8          & 79.3          & 89.7          \\
7                       & \textbf{}   & \textbf{}   & \checkmark      & 1,2(e)                                & \checkmark   & {\ul 97.1}    & {\ul 93.2}    & \textbf{80.5} & \textbf{90.3} \\
8                       &             &             & \checkmark      & 1,2(e)                                & -                           & 97.0          & 93.0          & {\ul 79.8}    & 89.9          \\
9                       &             &             & \checkmark      & 1,2(e),3                              & \checkmark   & 97.0            & 92.7          & 79.7          & 89.8          \\
10                      &             &             & \checkmark      & 1,2(e),3,4                            & \checkmark   & 96.8          & 92.9          & 79.7          & 89.8          \\
\hline
\end{tabular}
\end{table*}

To demonstrate the effectiveness and analyze the contributions of each component of our method, we have performed several groups of rigorous ablation experiments on CASIA-B in terms of different settings.
\subsubsection{Analysis of bilateral motion-oriented features}
The first three rows of Table~\ref{table6} report the impact of the bilateral motion-oriented features. By comparing the first two rows, the performance is improved by more than 1\% overall, and especially on the CL subset, the accuracy is enhanced by 1.5\%. Obviously,  motion-oriented features can efficiently represent the motion at the feature level, which complement the static information captured by the CNN. The difference of settings between the second and the third row is that the bilateral motion is considered, and the bilateral one yields the better result, 97.3\% on the NM subset and over 90\% on the mean accuracy. Accordingly, the findings support that taking advantage of bilateral information in $G_d$ is conducive to leverage motion features.
\subsubsection{Analysis of multi-scale temporal features}
The proposed method can produce a set of bilateral motion-oriented features at several temporal resolutions, as mentioned in Section \ref{3.3}. In the experiments, the model is investigated on four settings of different scales, as shown in Table~\ref{table6} where ‘-‘ denotes there is only one scale by default. As the number of scales increases, the features comprise more refined temporal information and cover a longer term consisting of more frames. Besides, we offer two kinds of ways to calculate the features at the second scale: a stacking approach (s) represented by Eq. \ref{eq11} and an embedded approach (e) represented by Eq. \ref{eq12}. We can observe that: (1) The 6th and the 7th row reveal the result from the two ways, and the embedded one achieves superior performance. This might stem from the deeper network in the stacking model, which involves too many convolutional layers with 1×1 kernels to train well. Hereby, we adopt the embedded approach as the representation at the second scale in the rest of the experiments. (2) The comparison between the 5th and the 7th row exhibits that the result of adding the second-order information surpasses that of employing the single features. Especially in the case of changing clothes, the improvement reaches 0.8\%, and there is only a slight drop from the optimal results of other settings on the NM subset and the BG subset. This is because refined spatiotemporal features can better reflect the person's identity information in complex conditions. (3) The results shown on the 5th, 7th, 9th, and 10th rows report the performance on four groups of features from different scales. The motion-oriented features from the first two resolutions achieve the top accuracy, 90.3\%, since leveraging information from two scales can provide a more comprehensive depiction of motion context. Moreover, it can be seen that the 9th and 10th rows do not achieve higher results than the 7th row. There is no further improvement in performance by adding richer features from more scales. The main reason might be the fine-grained features become too abstract to represent the identity.
\subsubsection{Analysis of correction}
In Table~\ref{table6}, we set three groups of comparison to evaluate the correction block: the second and the fourth rows, the third and the fifth rows, as well as the 7th and the 8th rows. The experiments are carried out based on motion-oriented features, bilateral motion-oriented features, and multi-scale bilateral motion-oriented features, individually. Obviously, multi-scale bilateral motion-oriented features with the correction yield the highest accuracy, and the correction block boosts the result by 0.4\% of the mean, and 0.7\% of the CL subset. These finding support that the correction block can avoid the negative impact of segmentation noise to some extent. Besides, we notice that there are no effects on the performance of features at the single scale, as shown in the comparison between the second and the fourth rows, the third and the fifth row. This may be explained by the CNN which can filter out some noise preliminarily. However, the multi-scale features are more sensitive and susceptible to the incorrect segmentation, since the refined temporal information is highly coupled with such noise at the edge of silhouettes. Consequently, the evident improvement caused by the correction block occurs with the multi-scale features, rather than the single-scale features. 
\subsubsection{Analysis of loss}

\begin{table}[!t]
\caption{Ablation experiments conducted on CASIA-B in terms of various settings on loss.\label{table7}}
\centering
\begin{tabular}{|c|c|c|c|c|c|c|c|}
\hline
\multirow{2}{*}{Number} & \multirow{2}{*}{LOSS}        & \multicolumn{4}{c|}{RESULTS} \\
\cline{3-6}
                        &                              & NM    & BG    & CL   & Mean \\
\hline
1                       & CE                           & 90.8  & 84.4  & 60.4 & 78.5 \\
2                       & Focal Loss ($\lambda $=2)             & 91.1  & 84.8  & 58.5 & 78.1 \\
3                       & Triplet                      & 96.5  & 92.2  & 78.4 & 89.0 \\
4                       & Triplet + CE                 & 96.9  & 93.1  & 79.4 & 89.8 \\
5                       & Triplet + Focal   Loss ($\lambda $=1) & {\ul 97.0}  & 92.7  & 80.0 & 89.9 \\
6                       & Triplet + Focal   Loss ($\lambda $=2) & \textbf{97.1}  & \textbf{93.2}  & \textbf{80.5} & \textbf{90.3} \\
7                       & Triplet + Focal   Loss ($\lambda $=5) & 96.8  & {\ul 93.1}  & {\ul 80.3} & {\ul 90.1} \\
\hline
\end{tabular}
\end{table}

We combine the triplet loss and the focal loss in this paper, and the effectiveness of different loss combinations is examined in this section. Table~\ref{table7} exhibits the result under several loss settings. (1) When the CE loss or the focal loss is independently deployed, the mean accuracy is not ideal, less than 79\% indeed. This is due to the fact that the identification phase of gait recognition is based on the distance metric, and the CE loss or the focal loss is fundamentally used for classification. Compared to the triplet loss, they hardly guide the distinctive feature map that lies in the inner layer of the network to represent the identities directly. (2) Jointly employing these two kinds of losses boost the performance of the network, because the CE or the focal loss is conductive to assist the triplet loss to discover a more discriminative gait space. (3) CE can be regarded as a special focal loss in which $\lambda $ is 0. Under this circumstance, the easy samples have the same importance as the difficult ones. This is the reason why applying the focal loss can achieve better results, 90.3\%, improved by 0.5\%, as shown in the fourth and the sixth rows. (4) A too large $\lambda $ forces the network to focus too much on difficult samples, which would also undermine the performance, as shown in the 7th row. Through the experiments, we find that the balance of learning can be reached to yield the best results when $\lambda $ is 2.

\section{Conclusion}
Unlike other gait recognition studies, the purpose of this research is to depict the gait motion pattern at the level of features in 2D-CNNs, which is direct and intrinsic for gait recognition. Firstly, the bilateral motion-oriented block, inspired by the definition of optical flow, is proposed to describe the motion context. Second, to represent the motion fully, such features are generalized to multiple temporal scales as refinement. Then, a correction block is designed to mitigate the adverse effect of segmentation noise. Lastly, the experimental results confirm the proposed method can boost the performance and yield the high accuracy compared with other current methods. Furthermore, as the motion dynamics extraction method is of importance for video understanding tasks, this method could be adopted in other fields, like emotion recognition, in the future.

\end{document}